\documentclass{article}
\pdfoutput=1
\usepackage[preprint]{neurips_2024}
\setcitestyle{authoryear,round,semicolon}
\usepackage[utf8]{inputenc}
\usepackage[T1]{fontenc}
\usepackage[hidelinks,hypertexnames=false]{hyperref}
\usepackage{float}   
\usepackage{url}
\usepackage{booktabs}
\usepackage{amsfonts}
\usepackage{amsmath}
\usepackage{amssymb}
\usepackage{nicefrac}
\usepackage{microtype}
\usepackage{graphicx}
\usepackage{xcolor}
\usepackage{enumitem}
\usepackage{pgfplots}
\pgfplotsset{compat=1.17}
\usetikzlibrary{arrows.meta, positioning, fit, calc}
\graphicspath{{figs/}}

\newcommand{\RfinalChanceX}{9.6}          
\newcommand{\RfinalPresentX}{2.3}         
\newcommand{\Chance}{0.040}                 
\newcommand{\SepLast}{0.317}                
\newcommand{\SepRho}{1.000}                 
\newcommand{\SepPexact}{4.0\times 10^{-4}}              
\newcommand{\SepN}{7}                   
\newcommand{\Present}{0.125}                
\newcommand{\RepairCtrlK}{18}            
\newcommand{\RepairCtrlN}{18}            
\newcommand{\AccRho}{+1.000}                 
\newcommand{\AccP}{4.0\times 10^{-4}}                   
\newcommand{\AccGapRho}{+0.964}              
\newcommand{\RdCmRho}{-0.893}                
\newcommand{\RdCmP}{0.01}                  
\newcommand{\AlignRho}{+0.429}               
\newcommand{\AlignP}{0.35}                 
\newcommand{\RlFrozenFirst}{0.864}          
\newcommand{\RlFrozenLast}{0.714}           
\newcommand{\RlRefitFirst}{0.864}           
\newcommand{\RlRefitLast}{0.792}            
\newcommand{\RlFrozenRho}{-1.000}            
\newcommand{\RlFrozenP}{4.0\times 10^{-4}}              
\newcommand{\RlRefitRho}{-0.893}             
\newcommand{\RlRefitP}{0.012}               
\newcommand{\RlFrozenLoss}{0.150}           
\newcommand{\RlRefitLoss}{0.072}            
\newcommand{\CuseN}{7}                  
\newcommand{\CuseFirst}{0.019}              
\newcommand{\CuseLast}{0.668}               
\newcommand{\CuseRho}{+0.964}                
\newcommand{\CuseP}{2.8\times 10^{-3}}                  
\newcommand{\CuseRandLast}{0.077}           
\newcommand{\CuseBigN}{7}               
\newcommand{\CuseBigFirst}{0.158}           
\newcommand{\CuseBigLast}{0.511}            
\newcommand{\CuseBigRho}{+0.929}             
\newcommand{\CuseBigP}{6.7\times 10^{-3}}               
\newcommand{\DevBigBehavFirst}{0.075}       
\newcommand{\DevBigBehavLast}{0.417}        
\newcommand{\DevSmallProbeLo}{0.058}        
\newcommand{\DevSmallProbeHi}{0.150}        
\newcommand{\DevSmallProbeRho}{-0.018}       
\newcommand{\LadderGapRhoLo}{+0.83}         
\newcommand{\LadderGapRhoHi}{+0.98}         
\newcommand{\LadNSixteenM}{9}           
\newcommand{\LadNFourTenM}{10}           
\newcommand{\LadNOneB}{14}               
\newcommand{\LadNSixNineB}{7}           
\newcommand{\LadNOneFourB}{7}           
\newcommand{\LadderGapSigTotal}{5}      
\newcommand{\OutgrownOneBRho}{-0.90}        
\newcommand{\OutgrownOneBP}{2.0\times 10^{-5}}          
\newcommand{\OutgrownBreakRho}{+0.36}       
\newcommand{\OutgrownBreakP}{0.44}         
\newcommand{\LadderCensored}{2.8B}         
\newcommand{\LadGapRhoSixteenM}{+0.98}      
\newcommand{\LadMarRhoSixteenM}{-0.86}      
\newcommand{\LadAccRhoSixteenM}{-0.24}      
\newcommand{\LadGapRhoFourTenM}{+0.83}      
\newcommand{\LadMarRhoFourTenM}{-0.81}      
\newcommand{\LadAccRhoFourTenM}{+0.47}      
\newcommand{\LadGapRhoOneB}{+0.86}          
\newcommand{\LadMarRhoOneB}{-0.90}          
\newcommand{\LadAccRhoOneB}{+0.89}          
\newcommand{\LadGapRhoSixNineB}{+0.95}      
\newcommand{\LadMarRhoSixNineB}{+0.36}      
\newcommand{\LadAccRhoSixNineB}{+0.96}      
\newcommand{\LadStrongScales}{3}        
\newcommand{\LadderPresent}{0.167}          
\newcommand{\OlmoOneBRho}{+0.851}            
\newcommand{\OlmoOneBP}{8.0\times 10^{-5}}              
\newcommand{\OlmoOneBXK}{2.33}             
\newcommand{\OlmoOneBN}{14}              
\newcommand{\OlmoOneBClears}{7}         
\newcommand{\OlmoOneBClearsN}{7}        
\newcommand{\OlmoSevenBRho}{+0.029}          
\newcommand{\OlmoSevenBP}{1.00}            
\newcommand{\OlmoSevenBXK}{1.20}           
\newcommand{\OlmoSevenBN}{6}            
\newcommand{\OlmoThirteenBRho}{+0.680}       
\newcommand{\OlmoThirteenBP}{3.0\times 10^{-3}}         
\newcommand{\OlmoThirteenBXK}{1.58}        
\newcommand{\OlmoThirteenBN}{17}         
\newcommand{\OlmoThirteenBClears}{8}    
\newcommand{\OlmoThirteenBClearsN}{9}   
\newcommand{\CruxN}{4}                  
\newcommand{\CruxEstab}{0}              
\newcommand{\CruxInputXcLo}{12.7}          
\newcommand{\CruxInputXcHi}{16.0}          
\newcommand{\CruxOutputXcLo}{2.7}         
\newcommand{\CruxOutputXcHi}{5.1}         
\newcommand{\CruxFailLo}{7}             
\newcommand{\CruxFailHi}{12}             
\newcommand{\CruxFailFloor}{40}          
\newcommand{\BoxDeclLo}{0.54}              
\newcommand{\BoxDeclHi}{0.87}              
\newcommand{\BoxTrackLo}{0.00}             
\newcommand{\BoxTrackHi}{0.29}             
\newcommand{\RcModels}{6}               
\newcommand{\RcXfairLo}{5.8}              
\newcommand{\RcXfairHi}{9.3}              
\newcommand{\RcRankTwoLo}{0.29}            
\newcommand{\RcRankTwoHi}{0.41}            
\newcommand{\RcBeatsOut}{0}             
\newcommand{\RcBindRec}{5.5}              
\newcommand{\RcKill}{not triggered}                 
\newcommand{\RcDistinctN}{4}            
\newcommand{\RcDistinctEstab}{3}        
\newcommand{\RcGapXcLo}{2.8}              
\newcommand{\RcGapXcHi}{3.7}              
\newcommand{\RcAccLo}{0.70}                
\newcommand{\RcAccHi}{0.75}                
\newcommand{\RcProbeChance}{0.10}          
\newcommand{\RcDecClean}{0.49}             
\newcommand{\RcWrongClean}{0.33}           
\newcommand{\RcBehavChance}{0.31}          
\newcommand{\RcGapPresentXLo}{0.9}        
\newcommand{\RcGapPresentXHi}{1.2}        
\newcommand{\RcStraddleEstab}{4}        
\newcommand{\RcStraddleN}{4}            
\newcommand{\RcRecBindMean}{0.25}          
\newcommand{\RcRecLastMean}{0.12}          
\newcommand{\CorpusN}{11,583}                
\newcommand{\CorpusRecallFail}{0.68}       
\newcommand{\CorpusRecallPass}{0.81}       
\newcommand{\CorpusCovFail}{0.55}          
\newcommand{\CorpusCovPass}{0.74}          
\newcommand{\CorpusSawNeverEdited}{27}   
\newcommand{\GossNeverEditFixed}{28}     
\newcommand{\GossPaccWrong}{0.73}          
\newcommand{\SftFailBase}{194}            
\newcommand{\SftFailSFT}{58}             
\newcommand{\SftFailDPO}{49}             
\newcommand{\SftFailRLVR}{48}            
\newcommand{\SftProbeBase}{0.787}           
\newcommand{\SftProbeSFT}{0.953}            
\newcommand{\SftSpearFirst}{0.367}          
\newcommand{\SftSpearLast}{0.667}           
\newcommand{\RlStepN}{5}                
\newcommand{\RlStepSpan}{2,400}             
\newcommand{\RlStepRecallFirst}{0.590}      
\newcommand{\RlStepRecallLast}{0.510}       
\newcommand{\DevTrnAccFirst}{0.261}         
\newcommand{\DevTrnAccLast}{0.972}          
\newcommand{\DevTrnBehavLast}{0.610}        
\newcommand{\DevTrnProbeOnFail}{0.953}      
\newcommand{\DevTrnBaseline}{0.167}         
\newcommand{\DevTrnGapSlope}{+0.179}         
\newcommand{\DevTrnGapSlopeCILo}{0.126}     
\newcommand{\DevTrnGapSlopeCIHi}{0.279}     
\newcommand{\DevTrnSmallSlope}{+0.058}       
\newcommand{\DevTrnSmallSlopeCILo}{-0.051}   
\newcommand{\DevTrnSmallSlopeCIHi}{0.143}   
\newcommand{\OnsetLag}{+0.061}               
\newcommand{\OnsetLagCILo}{-0.100}           
\newcommand{\OnsetLagCIHi}{0.212}           
\newcommand{\paccw}{\mathrm{pAcc}\!\mid\!\mathrm{wrong}}
\title{Decodable In-Context State and Model Output Across Training}

\author{%
  Manas Venkata Sai Ravulapalli\\
  Efficient Computation Inc.\\
  \texttt{manas@perseus.so}
  \And
  Samrath Chadha\\
  Efficient Computation Inc.\\
  \texttt{samrath@perseus.so}
}

\begin{document}
\maketitle

\begin{abstract}
Prior work established that a probe can decode an in-context binding on model errors and that probe-guided steering can repair some of them \citep{legible2609}. We follow probe accuracy, model output, and steering response across public pretraining and post-training checkpoints. Probe accuracy rises during Pythia pretraining, while probe-guided steering moves from negligible all-trial benefit to a larger benefit at two model sizes. Saved scores distinguish probe-correct errors with low and above-uniform model probability for the correct candidate. Oracle-target steering already repairs many early errors, but saved aggregates cannot separate target quality from intervention sensitivity. A held-out comparison of decoders trained on the final state or candidate logits finds no detected final-state advantage on late-checkpoint model errors. An information-theoretic counterexample explains why decodability on errors alone cannot establish discarded output information. The connection to downstream omissions remains open.
\end{abstract}

\section{Introduction}
An earlier study established that a language model can choose the wrong in-context binding even when a separately trained probe decodes the correct one from its hidden state \citep{legible2609}. It also showed that a probe-derived direction can change that choice. We ask how decodability, ordinary output, and intervention response change across checkpoints. The earlier results do not identify where the model's normal computation fails.

\begin{figure}[t]
  \centering
  \includegraphics[width=0.97\linewidth]{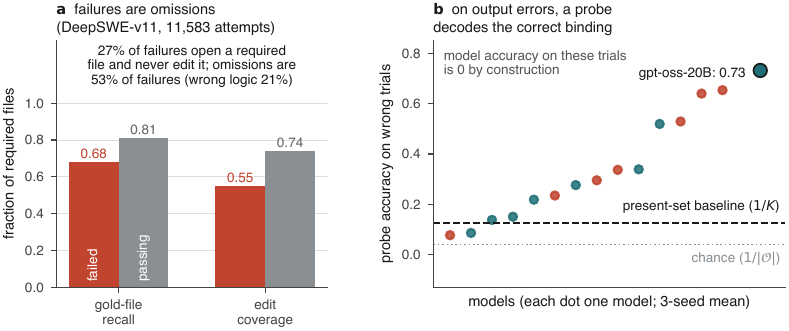}
  \caption{\textbf{A downstream motivation and a controlled binding result.} \emph{(a)} Failing repository-fix attempts find gold files nearly as
  well as passing ones (recall $\CorpusRecallFail$ vs $\CorpusRecallPass$) but edit far fewer ($\CorpusCovFail$
  vs $\CorpusCovPass$); $\CorpusSawNeverEdited\%$ of failures open a required file and never edit it (DeepSWE-v11,
  $\CorpusN$ attempts). These corpus statistics come from a transcribed summary; the raw source is not in the release.
  \emph{(b)} Probed on controlled binding, a separate probe often decodes the correct obligation on trials with an incorrect model output, following the protocol of \citet{legible2609};
  gpt-oss-20B has $\paccw=\GossPaccWrong$ (model accuracy on these trials is $0$ by construction).
  The repository episodes and binding trials are separate experiments; this figure does not establish a causal link between them.}
  \label{fig:puzzle}
\end{figure}

The motivating case came from an open model used for long-horizon repository work. In a fixed-sampling gpt-oss-20B run, $\GossNeverEditFixed\%$ of episodes ended without an edit. That observation motivates a test of represented state and output; it does not show that the omitted edit had the same cause as a binding error. In earlier controlled binding experiments, a separate probe decoded the correct binding on some incorrect model responses, above the present-set baseline (Figure~\ref{fig:puzzle}; \citealp{legible2609}). The probe is an external decoder, not the model's own readout.

We study the distinction on \emph{binding}, which associates an entity with an attribute and holds the association intact while other tokens compete for the same representational resources. Prior work identifies binding-identity structure in the models it studies \citep{binding2310}.
In the application of coding agents, an agent must keep \texttt{lock}$\rightarrow$\texttt{balance} apart from \texttt{mutex}$\rightarrow$\texttt{queue}, or a planner must track which subtask belongs to which actor; both of these settings relate to binding.

We find different checkpoint trajectories for probe accuracy, output accuracy, and probe-guided repair. The fixed-layer Pythia probes clear the present-set baseline at several larger scales (Section~\ref{sec:access}). A held-out comparison with candidate-logit decoders does not support a final-state information advantage on sampled late-checkpoint errors (Section~\ref{sec:heldout}). OLMo-2 summaries suggest a second family but lack a committed generator. An intermediate-layer comparison of gradient-derived and class-mean subspaces changes at Pythia-1B and 1.4B but reverses direction at 6.9B; it does not establish a final-readout mechanism (Section~\ref{sec:readout}). Probe-guided steering becomes more effective later in pretraining (Section~\ref{sec:repair}). In post-training, output behavior and frozen-probe accuracy change differently from accuracy of a probe refitted at each checkpoint (Section~\ref{sec:posttrain}). Whether these patterns explain omitted downstream actions remains a separate question.

\section{Setup}\label{sec:setup}
\paragraph{Task.} Each trial declares $K$ bindings ``\texttt{$e_i$: $o_i$}'' (entity $e_i$, single-token
obligation $o_i$, drawn from disjoint pools), then an interference block of $D$ distractor tokens, then the
query ``\texttt{The task for $e_j$ is:}'' for a random $j$. The model is
correct if and only if its top obligation-token logit is $o_j$. Entities and obligations are resampled independently on every trial, so the
identity of the queried entity carries no information about its obligation. Chance over the full pool is
$1/|\mathcal{O}|=0.04$; the \emph{present-set baseline}, the chance of naming one of the $K$ obligations actually in
context, is $1/K$. We report against $1/K$ throughout, and give it beside any looser $1/|\mathcal{O}|$ multiplier.

\paragraph{Probe protocols.} We fit logistic-regression probes on the residual stream at the query's last position to decode the queried obligation $o_j$. The Pythia geometry sweeps use a fixed middle layer, with probe regularization selected on training-fold cross-validation and accuracy reported on a disjoint test fold. A separate developmental implementation chooses a layer on a validation fold and evaluates on a test fold. The older Pythia phase-plane sweep chose its best layer using test accuracy; we exclude that plot and do not use it to support the developmental claim. The OLMo-2 checkpoint sweep has committed summary rows but no committed probe generator, so its layer-selection protocol cannot be audited here. The quantity of interest in the model-error battery is
$\paccw$, the probe's accuracy \emph{on trials where the model's output is wrong}. On those trials
the model's own accuracy is zero by construction. Probe accuracy above chance shows that this classifier can recover information associated with the binding. It does not show that the model's logits lack that information or identify the cause of the incorrect choice.

\paragraph{Definitions and baselines.} We report $\paccw$ and probe accuracy at a named layer. The Pythia geometry sweeps below use a fixed middle layer and compare probe accuracy in the full state with accuracy after projection onto gradient-derived or class-mean subspaces at that layer. The gradients are of a gold-label output margin through the remaining network; their principal directions are not the principal directions of the final output map. A separate depth analysis probes the final state without layer selection.

The descriptive probe gap is $\paccw$ minus the present-set baseline $1/K$. The subspace contrast is gradient-subspace probe accuracy minus class-mean-subspace probe accuracy at the same $k$. Its sign compares two probes at one intermediate layer. It is not a measure of information lost by the model's final output map.

Each experiment reports three baselines: the obligation-pool chance $1/|\mathcal{O}|=\Chance$, the value-vocabulary chance $1/|V|=\RcProbeChance$ (used on executed code in Section~\ref{sec:boundary}), and the present-set baseline $1/K$. The present-set value is $\Present$ at $K{=}8$ and $\LadderPresent$ at $K{=}6$, and it is our primary baseline. The symbol $\ell^{\ast}$ denotes the validation-selected layer. Lowercase $k$ is a subspace dimension, uppercase $K$ is the binding load, and $\rho$ is Spearman correlation against pretraining step unless another axis is named.

The \emph{query site} is the final query token, where the retrieved obligation is available. A \emph{frozen}
probe is fitted once and applied unchanged. A \emph{refit} probe is fitted at each checkpoint and follows
changes in the representation or probe compatibility.

\paragraph{Models.} The pretraining sweeps use the public checkpoints of the Pythia ladder (160M, 410M, 1B,
1.4B, 6.9B; 2.8B is censored for duplicate weights) and of OLMo-2 (1B, 7B, 13B). The post-training
measurements use the OLMo-2-7B base/SFT/DPO/Instruct ladder and the per-step branches of
\texttt{Olmo-3-7B-RL-Zero-Code}. We do not fine-tune models; measurements use frozen public checkpoints.

\subsection{Decodability and output are different tests}\label{sec:identification}
Let $Z$ be the correct binding, $R$ the final query-position state immediately before the output map, $S$ the vector of candidate logits, and $\hat Z=\arg\max S$. The model computes $S$ from $R$, so data processing gives
\begin{equation}\label{eq:information-order}
  \mathsf H(Z\mid R)\ \leq\ \mathsf H(Z\mid S)\ \leq\ \mathsf H(Z\mid\hat Z),
  \qquad I(Z;R\mid S)=\mathsf H(Z\mid S)-\mathsf H(Z\mid R).
\end{equation}
Here $\mathsf H$ denotes conditional entropy on the trial distribution. The first difference measures information absent from the candidate logits in an ideal-decoder sense. Comparing a fitted probe with $\hat Z$ does not estimate it. For example, let $Z$ be a uniform binary label, $R=Z$, and the two candidate logits be $S=(0,Z-2)$. The model always predicts class zero, but the second logit determines $Z$ exactly. On every error a probe of $R$ is correct, although a decoder of $S$ is equally capable. This counterexample separates a wrong argmax from loss of information in the scores. Fitted decoder comparisons remain subject to probe capacity, calibration, and held-out sampling error; the identity alone does not identify a causal route.

The identity gives a falsifiable test. If the candidate scores retain all label information available in $R$, Bayes-optimal decoders of $S$ and $R$ have equal expected log loss. A reproducible state-decoder advantage under a well-controlled fitted comparison would support information loss between $R$ and $S$, subject to decoder capacity and sample size. The comparison must use matched trials, including the model-error subset, and an independent prompt template. Even an advantage would not show that the missing information causes downstream action omissions.

\subsection{Held-out decoder comparison}\label{sec:heldout}
We ran an exploratory version of this test at two Pythia-1.4B checkpoints. The battery has $K=6$ bindings, 25 candidate obligations, and separate declaration and query templates for training, validation, and test. We fitted multinomial logistic decoders to the final query-position state and to the same trials' candidate logits. Each decoder uses training-fold standardization; its regularization is selected on the validation template. We compare paired test log loss, with a bootstrap over test episodes, and report the subset where the unmodified model's top candidate is wrong. Negative values in Table~\ref{tab:heldout} favor the candidate-logit decoder.

\begin{table}[t]
\centering\small
\caption{\textbf{Held-out final-state versus candidate-logit decoding on native model errors.} $\Delta$ is candidate-logit-decoder log loss minus final-state-decoder log loss. The interval resamples test episodes and does not cover training seeds or prompt templates. One model seed and one held-out test template were evaluated.}
\label{tab:heldout}
\begin{tabular}{lrrrr}
\toprule
Pythia-1.4B step & Wrong trials & $\Delta$ log loss & Bootstrap 95\% interval & Accuracy: state / logits \\
\midrule
1{,}000 & 232 & $-1.811$ & $[-2.053,-1.564]$ & $0.047/0.026$ \\
143{,}000 & 193 & $-0.007$ & $[-0.179,0.162]$ & $0.140/0.181$ \\
\bottomrule
\end{tabular}
\end{table}

At the late checkpoint, the interval includes zero, so this fitted comparison does not detect a final-state advantage on model errors. On all test trials, the late-checkpoint difference is $0.096$ with a paired interval of $[-0.051,0.238]$ ($n=250$). The early-checkpoint candidate-logit decoder has lower log loss on model errors. Neither result estimates the Bayes-optimal information difference in Equation~\ref{eq:information-order}: decoder capacity, 20 training examples per class, and template shift limit that inference. The observed probe-versus-argmax gap therefore does not establish a final-readout bottleneck.

The raw candidate scores give a simpler decision baseline on these held-out trials. On native errors, the correct candidate ranks second in $12/232$ cases at step 1{,}000 and $43/193$ at step 143{,}000. The median winner-minus-correct logit gap falls from $2.935$ to $1.629$. Thus the correct candidate moves closer to the winning answer in this comparison, but most late errors still rank it below second. These are different error subsets at the two checkpoints; rank and margin do not measure information absent from the full score vector.

\section{Probe accuracy changes over pretraining}\label{sec:access}
Over pretraining, fixed-middle-layer probe accuracy rises at several Pythia scales. Accuracy after projection onto a gradient-derived subspace follows a different trajectory.

\paragraph{What changes among errors.} A saved Pythia-1.4B posterior battery crosses two tests on each wrong-answer trial: whether the probe's top candidate is correct ($P^+$), and whether the model assigns the correct candidate at least uniform probability among the $K=6$ present candidates ($S^+$). Table~\ref{tab:error-groups} shows counts rather than conditioning away the falling error rate. The probe-correct, low-score group grows from $64$ to $147$ trials, while the probe-correct, above-uniform group grows from $11$ to $78$. A low score does not mean the logits lack label information: a decoder may still use their full pattern. Probe layers were selected separately at each checkpoint, and the saved rows lack trial identifiers. The table describes group composition; it does not establish a paired transition or a cause of the errors.

\begin{table}[t]
\centering\small
\caption{\textbf{Operational error groups in saved Pythia-1.4B posterior batteries.} Each row contains $400$ trials with $K=6$ present candidates. $P^+$ means probe top-one correct; $S^+$ means model probability of the correct candidate at least $1/K$. Superscript minus denotes the complementary test. The four groups partition native model errors. Probe layer is selected separately for each checkpoint; prompt identities are not saved.}
\label{tab:error-groups}
\begin{tabular}{rrrrrr}
\toprule
Step & Errors & $P^+S^+$ & $P^+S^-$ & $P^-S^+$ & $P^-S^-$ \\
\midrule
1{,}000 & 330 & 11 & 64 & 34 & 221 \\
8{,}000 & 321 & 10 & 64 & 37 & 210 \\
32{,}000 & 306 & 22 & 134 & 24 & 126 \\
143{,}000 & 272 & 78 & 147 & 7 & 40 \\
\bottomrule
\end{tabular}
\end{table}

\begin{figure}[t]
  \centering
  \includegraphics[width=0.98\linewidth]{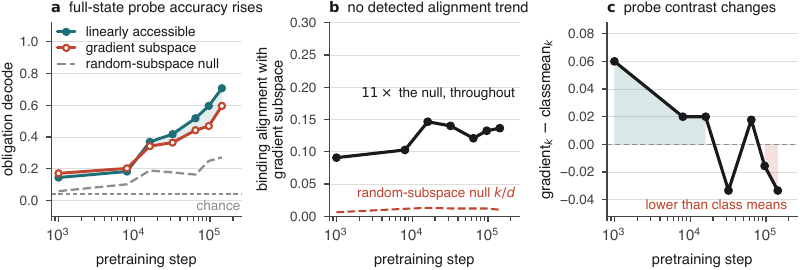}
  \caption{\textbf{Fixed-middle-layer probe geometry} (Pythia-1.4B). \emph{(a)} Full-state binding probe accuracy rises
  ($\rho=\AccRho$, exact $p=\AccP$). \emph{(b)} Alignment of the gradient-derived subspace with binding class means has no detected trend ($\rho=\AlignRho$, $p=\AlignP$).
  \emph{(c)} Gradient-subspace probe accuracy minus class-mean-subspace probe accuracy falls ($\rho=\RdCmRho$, exact $p=\RdCmP$).
  The gradient-derived subspace comes from a gold-label output margin through the remaining network. Neither projection isolates the model's natural readout.}
  \label{fig:access}
\end{figure}

\subsection{Across the Pythia ladder}
 At Pythia-1.4B the
binding becomes far more linearly accessible over pretraining ($\rho=\AccRho$, exact $p=\AccP$;
Figure~\ref{fig:access}a). Across the Pythia
ladder (Table~\ref{tab:ladder}), the full-state probe pulls away from the gradient-subspace probe at every usable
scale from 160M to 6.9B (the difference's $\rho$ runs from $\LadderGapRhoLo$ to $\LadderGapRhoHi$, exact
$p<0.05$ at all $\LadderGapSigTotal$). This contrast does not distinguish information growth from changing probe performance after projection.
Accessibility rises above the $1/K=\LadderPresent$ baseline at $\LadStrongScales$ scales: 1B
($\rho=\LadAccRhoOneB$), 1.4B ($\rho=\AccRho$), and 6.9B ($\rho=\LadAccRhoSixNineB$). At 160M and 410M it is
flat ($\rho=\LadAccRhoSixteenM$, $\LadAccRhoFourTenM$, n.s.) and ends at or below baseline. Those two scales do not support a claim of a reliably decodable binding.

\begin{table}[t]
\centering\small
\caption{\textbf{Pythia fixed-middle-layer probe geometry.} Per-scale Spearman $\rho$ vs pretraining step for
full-state probe accuracy, full-state minus gradient-subspace probe accuracy, and gradient-subspace minus class-mean-subspace probe accuracy;
$\dagger$ marks exact permutation $p<0.05$. Full-state accuracy clears $1/K=\LadderPresent$ at 1B/1.4B/6.9B.
The third contrast has an opposite point-estimate trend at 6.9B and is not significant there. These probe correlations do not test the final output map. Pythia-$\LadderCensored$ is censored (duplicate weights). $n$ counts
weight-distinct checkpoints; the 6.9B column's small $n$ limits its power.}
\label{tab:ladder}
\begin{tabular}{lcccc}
\toprule
Scale & $n$ & Full-state $\rho$ & Full$-$gradient $\rho$ & Gradient$-$class mean $\rho$ \\
\midrule
160M  & $\LadNSixteenM$ & $\LadAccRhoSixteenM$          & $\LadGapRhoSixteenM\,\dagger$ & $\LadMarRhoSixteenM\,\dagger$ \\
410M  & $\LadNFourTenM$ & $\LadAccRhoFourTenM$          & $\LadGapRhoFourTenM\,\dagger$ & $\LadMarRhoFourTenM\,\dagger$ \\
1B    & $\LadNOneB$     & $\LadAccRhoOneB\,\dagger$     & $\LadGapRhoOneB\,\dagger$     & $\LadMarRhoOneB\,\dagger$ \\
1.4B  & $\LadNOneFourB$ & $\AccRho\,\dagger$            & $\AccGapRho\,\dagger$         & $\RdCmRho\,\dagger$ \\
6.9B  & $\LadNSixNineB$ & $\LadAccRhoSixNineB\,\dagger$ & $\LadGapRhoSixNineB\,\dagger$ & $\LadMarRhoSixNineB$ \\
\bottomrule
\end{tabular}
\end{table}

\subsection{Scale dependence}
\begin{figure}[t]
  \centering
  \includegraphics[width=0.98\linewidth]{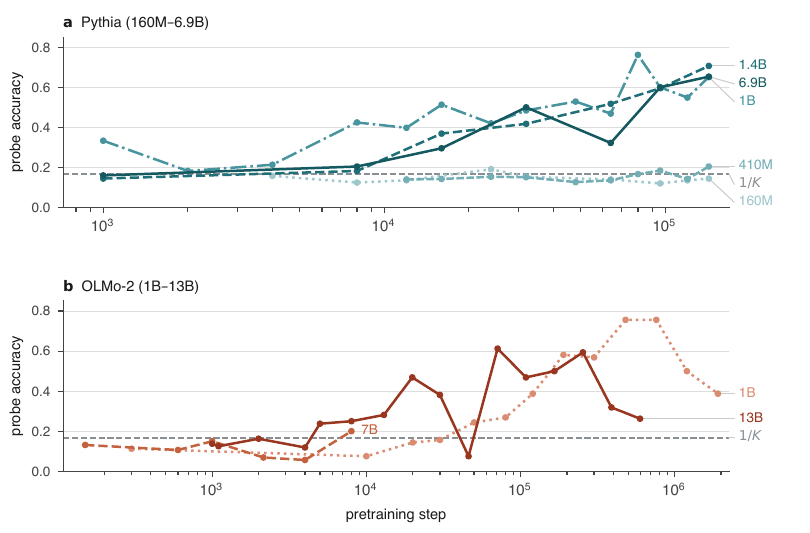}
  \caption{\textbf{Probe accuracy by model size.} Pythia fixed-middle-layer and OLMo recorded probe accuracy for the queried
  binding against pretraining step (log axis), all scales of a family overlaid; the dashed line is the $1/K$
  present-set baseline. Scales at and above 1B separate from the near-chance 160M/410M curves, whose
  paths end at or barely above the baseline; within the ${\ge}1$B group, final
  probe accuracy is similar (0.65--0.71). On OLMo-2 the 1B and 13B sweeps clear the baseline; the 7B sweep is truncated at 34B tokens. The OLMo probe generator is unavailable in the release.
  Line identity is carried by direct labels and dash patterns.}
  \label{fig:devscale}
\end{figure}

 Figure~\ref{fig:devscale} overlays
every scale of both families on shared axes. Scales at and above 1B separate from the near-chance 160M/410M
curves, whose paths end at or barely above the baseline; within the ${\ge}1$B group, final probe accuracy is
similar. The OLMo rows are descriptive because their probe generator is not committed.

\subsection{OLMo-2 summary rows}
The OLMo-2 summaries suggest a similar probe-accuracy rise. Their generator is absent from the release, so the layer-selection protocol and trial construction cannot be checked. On the recorded battery, OLMo-2-1B rises over $\OlmoOneBN$ checkpoints
(Spearman $\rho=\OlmoOneBRho$, exact $p=\OlmoOneBP$). It clears $1/K$ on
$\OlmoOneBClears/\OlmoOneBClearsN$ late checkpoints and ends at $\OlmoOneBXK\times$ baseline. OLMo-2-13B
also rises over $\OlmoThirteenBN$ checkpoints ($\rho=\OlmoThirteenBRho$, $p=\OlmoThirteenBP$), clearing the
baseline on $\OlmoThirteenBClears/\OlmoThirteenBClearsN$ late checkpoints and ending at
$\OlmoThirteenBXK\times$. The sparsely-sampled third size
(OLMo-2-7B, $\OlmoSevenBN$ checkpoints) is under-powered and does not reach significance
($\rho=\OlmoSevenBRho$, $p=\OlmoSevenBP$, $\OlmoSevenBXK\times$ baseline). These rows do not yet establish an independently reproducible cross-family result.
This OLMo-2 sweep tests a probe-accuracy trajectory. It does not replicate the Pythia subspace measurement or identify a readout mechanism.

\section{Depth and subspace comparisons}\label{sec:readout}
\subsection{Depth separation}
 At Pythia-1.4B, final-layer decoding reaches
$\RfinalPresentX\times$ the $1/K$ present-set baseline without layer selection, or $\RfinalChanceX\times$ the looser $1/|\mathcal{O}|$ vocabulary baseline.
The separation between the best and
final layers opens monotonically to $\SepLast$ (Spearman $\rho=\SepRho$, exact $p=\SepPexact$, $n=\SepN$).
The best-layer probe and final-layer probe have different checkpoint trajectories. This does not show that information disappears at the final layer or explain the model's output errors.

\subsection{Gradient and class-mean subspaces}
At Pythia-1.4B, alignment of the gradient-derived subspace with binding class means has no detected trend
($\rho=\AlignRho$, $p=\AlignP$, n.s.). The gradient-subspace probe falls relative to the class-mean-subspace probe
($\rho=\RdCmRho$, $p=\RdCmP$). The same fall is significant at Pythia-1B ($\rho=\OutgrownOneBRho$,
$p=\OutgrownOneBP$; Table~\ref{tab:ladder}). The trend is directionally negative at 160M/410M, where full-state probe accuracy is near chance. It does not replicate at \textbf{6.9B}, where the point estimate
is opposite and not significant ($\rho=\OutgrownBreakRho$, $p=\OutgrownBreakP$). This is a scale-dependent probe contrast at a fixed middle layer. Because the gradient directions are conditioned on the gold label and pass through the remaining network, their comparison with class means cannot identify the natural output mechanism.

\section{Repairability develops during pretraining}\label{sec:repair}
\subsection{Self-gated repair across checkpoints}\label{sec:cause}
 The developmental control extends the self-gated activation repair of \citet{legible2609} across checkpoints. At the read layer
$\ell^{\ast}$, we add a unit direction from the emitted class toward the class decoded by the probe scaled by
$\alpha\lVert h\rVert$ with $\alpha=0.5$, and re-run the remaining layers. The direction is zero whenever probe and
output agree, so no gold label enters.
On Pythia-1.4B checkpoints this repair
recovers $\CuseFirst$ of the failure set early and $\CuseLast$ late (Spearman $\rho=\CuseRho$, $p=\CuseP$,
$n=\CuseN$). A matched-norm wrong direction recovers $\CuseRandLast$ at the final checkpoint. This comparison supports a target-specific intervention effect. It does not show that the unmodified model uses the same direction.

Failure repair alone omits damage to correct answers. On all held-out trials at step 8{,}000, the self-gated intervention changes accuracy from $0.255$ to $0.238$ (Pythia-1.4B, $n=400$); at step 143{,}000, it changes accuracy from $0.323$ to $0.753$ on the same battery size. At Pythia-6.9B, the corresponding early change is $0.201$ to $0.190$ ($n=268$), and the late change is $0.321$ to $0.597$. These are descriptive checkpoint contrasts without paired trial-level uncertainty. The steering target comes from the probe, and the selected layer changes across checkpoints. Better target selection or changing intervention sensitivity could therefore explain the growing benefit.

\paragraph{What the repair arms identify.} The saved steering files compare an oracle target, the probe's decoded target, and a norm-matched random direction on the same baseline failures (Table~\ref{tab:repair-arms}). At Pythia-1.4B step 8{,}000, oracle-target steering repairs $239/298$ failures while probe-target steering repairs $56/298$. By step 143{,}000, the corresponding counts are $241/271$ and $174/271$. The large early oracle response rules out an account in which this intervention site is uniformly ineffective until late training. At Pythia-6.9B, both arms change substantially, so the same conclusion need not hold at that size. These aggregates do not isolate why probe-target repair grows.

\begin{table}[t]
\centering\small
\caption{\textbf{Repair of native failures at matched intervention strength.} Counts are repaired failures over the common failure set within each saved steering file. Oracle and probe arms use the same probe-weight direction construction with the gold or decoded target, respectively; the random direction is norm matched. Each arm uses $\alpha=0.5$. Layers differ across checkpoints.}
\label{tab:repair-arms}
\begin{tabular}{llrrrr}
\toprule
Model & Step & Failures & Oracle target & Probe target & Random \\
\midrule
Pythia-1.4B & 1{,}000 & 324 & 9 & 3 & 5 \\
 & 8{,}000 & 298 & 239 & 56 & 12 \\
 & 143{,}000 & 271 & 241 & 174 & 21 \\
Pythia-6.9B & 1{,}000 & 221 & 29 & 18 & 5 \\
 & 8{,}000 & 214 & 98 & 35 & 7 \\
 & 143{,}000 & 182 & 159 & 100 & 15 \\
\bottomrule
\end{tabular}
\end{table}

This ambiguity has an exact accounting form. On baseline errors $E$, let $Q$ indicate that the probe target equals the gold target, and let $Y_o$ and $Y_p$ indicate repair under the oracle and probe arms. The two interventions are identical when $Q=1$, so
\begin{equation}\label{eq:repair-decomposition}
\begin{aligned}
&\Pr(Y_o=1\mid E)-\Pr(Y_p=1\mid E)\\
&=\Pr(Q=0\mid E)\bigl[\Pr(Y_o=1\mid E,Q=0)\\
&\hspace{5.7em}-\Pr(Y_p=1\mid E,Q=0)\bigr].
\end{aligned}
\end{equation}
The identity is an accounting constraint, not a model of training. The aggregate arm rates omit both the fraction of failures with an incorrect probe target and the repair rates conditional on that event. Recording paired trial-level targets and outcomes would separate those terms; the present files do not.

\subsection{Replication at 6.9B}
At $6.9$B, repair on the failure set rises from $\CuseBigFirst$ to $\CuseBigLast$
($\rho=\CuseBigRho$, $p=\CuseBigP$, $n=\CuseBigN$). This repeats the
$\CuseFirst\!\to\!\CuseLast$ curve at 1.4B, while behaviour moves from
$\DevBigBehavFirst\to\DevBigBehavLast$. At Pythia-410M the decode
never develops ($\DevSmallProbeLo$--$\DevSmallProbeHi$, $\rho=\DevSmallProbeRho$, n.s.).
Probe accuracy remains near chance at this smaller scale, limiting the information available to the probe-guided intervention.

\subsection{Independent re-implementation}
 A second implementation shares no
analysis code and uses paired trials at every public Pythia-1.4B checkpoint. Its read
layer is fixed once on a selection fold. Accessibility rises from $\DevTrnAccFirst$ to $\DevTrnAccLast$, while
behaviour reaches $\DevTrnBehavLast$. The gap grows by $\DevTrnGapSlope$ per decade of training (bootstrap CI
$[\DevTrnGapSlopeCILo, \DevTrnGapSlopeCIHi]$). At the final checkpoint, the probe recovers the binding on
$\DevTrnProbeOnFail$ of failed trials against a $1/K=\DevTrnBaseline$ baseline. At Pythia-410M the slope is $\DevTrnSmallSlope$ (CI $[\DevTrnSmallSlopeCILo,
\DevTrnSmallSlopeCIHi]$, n.s.), which matches the scale scoping above. The pre-declared falsification test gives an onset lag of $\OnsetLag$ decades (CI $[\OnsetLagCILo,
\OnsetLagCIHi]$), so the trajectory does not support an onset-time difference between the curves. Onset is also left-censored because
accessibility exceeds baseline at the first checkpoint. The developmental claim therefore concerns asymptote
and slope.

\section{Post-training changes output and probe compatibility}\label{sec:posttrain}
We next compare public post-training checkpoints. These are observational contrasts across stages and do not isolate the effect of an individual training objective.

\subsection{SFT, DPO, and RLVR on OLMo-2-7B}
On OLMo-2-7B's four public checkpoints (base $\to$ SFT $\to$ DPO $\to$ Instruct/RLVR, the same instrument throughout), the largest measured change in probe accuracy occurs between base and SFT.
Recall failures fall $\SftFailBase\to\SftFailSFT$ per $400$ at SFT and then barely move ($\SftFailSFT\to\SftFailDPO\to\SftFailRLVR$). Probe decodability jumps $\SftProbeBase\to\SftProbeSFT$ and plateaus
(Figure~\ref{fig:sftrlvr}a). Probe and model posterior rankings become more correlated
(Spearman $\SftSpearFirst\to\SftSpearLast$). This comparison does not establish which stage formed the underlying representation.

\subsection{Per-step RL}
\begin{figure}[t]
  \centering
  \includegraphics[width=0.858\linewidth]{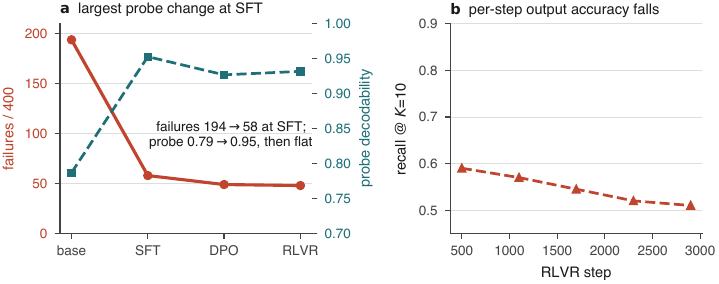}
  \caption{\textbf{Post-training checkpoint comparisons.} \emph{(a)} OLMo-2-7B across post-training: failures fall
  and probe decodability jumps at SFT, then plateau. \emph{(b)} per-step \texttt{Olmo-3-7B-RL-Zero-Code}:
  recall at $K{=}10$ falls across $\RlStepSpan$ RL steps; a probe on a held-out fold shows no trend
  (Figure~\ref{fig:rlsupp}). One family/seed per panel; $n=400$ in (a) and $n=300$ in (b).}
  \label{fig:sftrlvr}
\end{figure}

A per-step release (\texttt{Olmo-3-7B-RL-Zero-Code},
$\RlStepN$ collision-free step branches, GRPO on code) permits a within-branch trajectory. Across
$\RlStepSpan$ RL steps, recall at $K{=}10$ falls
$\RlStepRecallFirst\to\RlStepRecallLast$, while a held-out probe shows no trend
(Figure~\ref{fig:rlsupp}). By the late steps
the model no longer prefers the correct obligation at that load (Figure~\ref{fig:sftrlvr}c).
On this branch, output accuracy falls while refitted probe accuracy shows no detected trend. A non-significant probe trend is not proof that encoded information remains constant.

\subsection{Frozen and refitted probes}
Across GRPO steps on a coding reward, a probe frozen on the base model falls from $\RlFrozenFirst$
to $\RlFrozenLast$ (Spearman $\rho=\RlFrozenRho$, exact $p=\RlFrozenP$). A probe refit at each checkpoint falls
from $\RlRefitFirst$ to $\RlRefitLast$ ($\rho=\RlRefitRho$, $p=\RlRefitP$). The frozen probe loses
$\RlFrozenLoss$, compared with $\RlRefitLoss$ for the refitted one. The difference shows that a fixed probe becomes less compatible with later checkpoints. It does not decompose the change into geometric rotation and information loss.

\begin{figure}[t]
  \centering
  \includegraphics[width=0.86\linewidth]{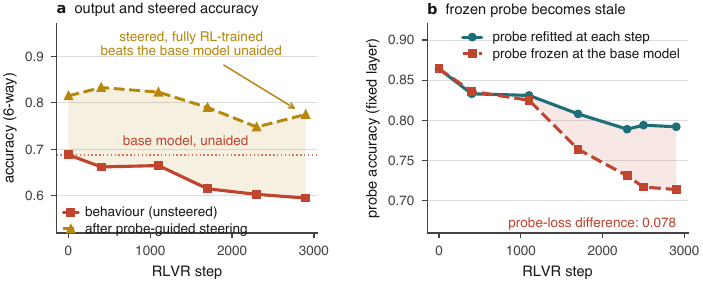}
  \caption{\textbf{Output, steering, and probe compatibility during code-reward training.} Output accuracy falls across the sampled steps; probe-guided steering changes output accuracy. A probe trained at the base checkpoint loses more accuracy than probes refitted at each step. The comparison shows probe staleness, not a change in the model's natural readout.}
  \label{fig:clocks}
\end{figure}

\section{Computed state: the boundary of the protocol}\label{sec:boundary}
On \emph{computed} state (a value a program produces rather than one written in the prompt), most tested instruments
do not establish probe accuracy above the present-set baseline on model errors. Surface presence and low error counts limit the tests below.
Table~\ref{tab:boundary} lists the three substrates and their controls. Between them they specify what a
computed-state instrument has to satisfy: a target the model produces internally, evaluable in behaviour, and
absent from the surface of the prompt.

\subsection{Tracked Python state (positive control)}\label{sec:gap}
 A probe-output separation appears on $\RcModels$ models tracking a variable's value through Python programs: on trials where the model
mis-predicts the value, a probe recovers it at $\RcXfairLo$--$\RcXfairHi\times$ the $1/K$ present-set baseline
and distinguishes binding from recency by $\RcBindRec\times$ on wrong trials. The recovered value is the
model's own \emph{second-ranked} token ($\RcRankTwoLo$--$\RcRankTwoHi$), and the probe never beats the model's
logits ($\RcBeatsOut$ of $\RcModels$). The task requires following a value through the program, so the target
is absent from the surface. The remaining subsections treat executed
code and computed outputs, where surface presence limits the protocol. This tracked-state result supplies the positive control.

\begin{table}[t]
\centering\small
\caption{\textbf{Computed-state substrates: what the probe reads, and why the protocol is power-limited
there.} Executed code straddles the present-set $1/K$; entity-tracking decodes the declaration, not the
tracked state; the pre-registered CRUXEval-O test is censored by model accuracy. A surface-literal control
names the shared confound. Decodes are in units of the stated chance baseline unless noted.}
\label{tab:boundary}
\footnotesize\setlength{\tabcolsep}{4pt}
\begin{tabular}{llp{3.0cm}}
\toprule
Substrate & Probe decode & Verdict \\
\midrule
Executed MBPP/HumanEval & $\RcGapXcLo$--$\RcGapXcHi\times\,1/|V|$; \ $\RcGapPresentXLo$--$\RcGapPresentXHi\times\,1/K$ & clears vocab, straddles $1/K$ ($\RcStraddleEstab/\RcStraddleN$) \\
Boxes entity-tracking & declaration $\BoxDeclLo$--$\BoxDeclHi$ vs.\ tracked $\BoxTrackLo$--$\BoxTrackHi$ & declaration-confounded \\
CRUXEval-O outputs & input $\CruxInputXcLo$--$\CruxInputXcHi\times$ vs.\ output $\CruxOutputXcLo$--$\CruxOutputXcHi\times$ & pre-registered \textsc{kill} ($\CruxEstab/\CruxN$) \\
\bottomrule
\end{tabular}
\end{table}

\subsection{Executed code}
 On $\RcDistinctN$ distinct
frontier code models (behavioural accuracy $\RcAccLo$--$\RcAccHi$), a probe of a variable's traced final value
clears the loose vocabulary baseline: $\RcDistinctEstab$ of $\RcDistinctN$ models beat the pre-registered
$2\times$ bar ($\RcGapXcLo$--$\RcGapXcHi\times\,1/|V|$), \textsc{kill} \emph{\RcKill}. But against the stricter
$1/K\approx\RcBehavChance$ present-set baseline it sits at $\RcGapPresentXLo$--$\RcGapPresentXHi\times$ and every
established evaluation's 95\% CI straddles it ($\RcStraddleEstab/\RcStraddleN$). The probe tracks the executed
value over recency ($\RcRecBindMean$ vs $\RcRecLastMean$).

\subsection{Wrong-content control on executed code}
 We pre-registered the causal-use control
(\texttt{PREREG\_realcode\_binding.md}) and re-ran it on \emph{real, executed} Python (MBPP/HumanEval canonical
solutions under \texttt{sys.settrace}, the target a variable's traced final value). The
decoded arm still beats wrong content on the cleanest model (deepseek, $\RcDecClean$ vs $\RcWrongClean$), but
wrong content itself repairs on every executed model, so the wrong-harms control that fires on all
$\RepairCtrlK/\RepairCtrlN$ synthetic runs does not replicate. On computed state the probe cannot
separate what the model holds from what the prompt already shows.

\subsection{Entity tracking and CRUXEval-O}
 On the Boxes
entity-tracking task a probe reads the initial \emph{declaration} ($\BoxDeclLo$--$\BoxDeclHi$) far better than
the \emph{tracked} state ($\BoxTrackLo$--$\BoxTrackHi$): what decodes is where the value was written, not what it
became. On pre-registered CRUXEval-O outputs the gap is established on no model ($\CruxEstab/\CruxN$,
\texttt{PREREG\_cruxeval.md}), and every arm is censored by model accuracy, with
$n_{\text{fail}}=\CruxFailLo$--$\CruxFailHi$ wrong trials against the pre-registered floor of
$\CruxFailFloor$, so the fired \textsc{kill} records an underpowered null rather than a decisive absence. A surface-literal
control shows what the probe still reads: at the answer slot the features decode the given \emph{input}
at $\CruxInputXcLo$--$\CruxInputXcHi\times$ chance but the \emph{computed output} at only $\CruxOutputXcLo$--$\CruxOutputXcHi\times$.

\subsection{Surface presence}
 The tested substrates leave distinct limits: some targets have a value token on the prompt surface, while
some confound-clean targets have too few distinct outputs (\texttt{[]}, \texttt{''}, booleans). The present
protocol does not separate decoding of computed state from surface evidence on those tests. Other instruments
built differently report probe-output gaps on counting and arithmetic targets
\citep{counting2605, arith2507, vlmcount2607}, thus the null is a statement about this protocol's power. The positive control bounds the scope from the other side: on tracked Python state,
which is not surface-written, a probe-output separation appears (Section~\ref{sec:gap}). Our supported scope is
declared bindings and this tracked-state control. A decisive computed-state test needs targets absent from the prompt surface, adequate error counts, and a calibrated candidate-logit baseline.

\section{Related work}\label{sec:related}
\paragraph{Hidden knowledge and the in-context gap.} \citet{orgad2410} and \citet{insideout2503}
report answers decodable internally but absent from output in factual question answering. In a separate
reasoning setting, \citet{detext2604} compare answer availability during a generated trace with forced
extraction. \citet{recall2510} argue that some internal signals track knowledge recall rather than truthfulness.
Our labels are new bindings supplied in the prompt, so the task does not test recall of a stored fact.

Our earlier study established the binding-error probe gap, query-entity control, failure detection, and non-oracle repair across fixed checkpoints \citep{legible2609}. Its Jacobian test did not find reduced output sensitivity to the binding direction on wrong trials. The present paper tests how probe accuracy and repair response change during training; it does not treat the gap or repair method as new.

Several concurrent studies examine the in-context gap. \citet{attndef2602} report a binding
dissociation and repair errors with an oracle target. \citet{lepori2602} study failures to deploy representations
defined in context. Related probe-output dissociations appear in repeated-token counting \citep{counting2605}
and causal yes/no tasks \citep{tonguetie2605}. \citet{agentmem2605} trace circuits involved in silent
agent-memory failures; those trajectories and our binding trials measure different tasks.
\citet{arith2507} use a probe-based error detector to select reasoning steps for re-prompting.

In sequential decision-making, \citet{schmied2504} find that a model can calculate the
highest-scoring bandit action in its written rationale yet choose another action. Their
``knowing--doing gap'' compares a generated calculation with a generated choice. Our
probe measures information decodable from a hidden state during training; neither
measurement alone identifies the model's natural route from information to action.

Latent-knowledge work also separates internal decodability from output: \citet{mallen2312} recover
answers from middle-layer probes in models fine-tuned to answer incorrectly. A probe alone does not
show how the unmodified model uses that information, and unsupervised elicitation can select an
unrelated prominent feature \citep{farquhar2312}. \citet{gurnee2607} use a context-averaged Jacobian
lens, concept swaps, and ablations to test when verbalizable representations influence computation.
Their J-space method measures a different object from our supervised binding probe; we do not infer
its workspace mechanism from probe accuracy or probe-guided steering.

\citet{kwon2607} study represented communicative intent and behavioral response across fixed models.
We study within-model trajectories across pretraining checkpoints and measure, at each checkpoint,
whether steering toward the probe's decoded target changes the output.

\paragraph{Representation and behaviour during training.} Internal structure can mature ahead of behaviour
\citep{repout2604}. \citet{billa2602} track probes across pretraining checkpoints and find task information
decodable before behaviour follows. We add an intervention and measure repairability over the same
developmental axis. Other trajectories are non-monotone: rules can collapse after emerging
\citep{ungrok2606}, and early semantic geometry can be transient \citep{transient2606}. The monotone rises in
Table~\ref{tab:ladder} are therefore empirical. Causal structure-before-behaviour results also appear on toy
tasks \citep{priors2607}.

\citet{devscale2605} measure whether next-token-prediction geometry sharpens with scale. Our axis is an
in-context binding outside the prediction target, so the two results concern different readouts.
\citet{staleness2606} independently find that
fine-tuning-style updates make frozen probes stale where quantization-style updates do not. This relates to the frozen-probe degradation observed here. Probe-based monitoring requires validation against behaviour after model updates.
Decodability alone does not establish use \citep{deconly2605, decgap2604}. Our developmental control addresses
one part of that distinction: probe-guided steering becomes more effective across sampled checkpoints. It does not establish the unmodified model's computation.

\section{Conclusion}\label{sec:limits}
The binding-error gap and probe-guided repair were established in earlier work \citep{legible2609}. Here fixed-middle-layer probe accuracy rises over several Pythia pretraining sweeps. Self-gated steering has little all-trial benefit at an early checkpoint but a larger benefit late in pretraining at two Pythia sizes. Saved error groups show that probe-correct failures can coexist with low model probability for the correct candidate; the threshold does not test information loss. Oracle-target steering already repairs many failures before probe-target steering does, but aggregate arms do not separate target quality from intervention sensitivity. A Pythia subspace contrast changes at 1B and 1.4B but has an opposite, non-significant trend at 6.9B. During code-reward training, output accuracy and frozen-probe accuracy fall on the sampled branch. OLMo-2 summary rows suggest a similar pretraining trajectory, but their probe generator is unavailable. The held-out decoder comparison does not detect a late final-state advantage over candidate logits on model errors. These tests do not identify why the unmodified model chooses the wrong answer or explain omitted repository edits. A causal intervention on a computed downstream task remains necessary for the latter claim.

\section*{Acknowledgements}
We would like to thank Kevin Li for his help with proofreading, for helping us to better write
the paper for readability, and for the suggestions that led to this paper.

\bibliographystyle{plainnat}

\begin{thebibliography}{99}
\setlength{\itemsep}{1pt}\setlength{\parskip}{0pt}\small

\bibitem[Billa(2026)]{billa2602}
J.~Billa. The geometric anatomy of capability acquisition in transformers. 2026. arXiv:2602.15997.

\bibitem[Cheang et al.(2025)]{recall2510}
C.~S. Cheang, H.~P. Chan, W.~Zhang, and Y.~Deng. Do LLMs really know what they don't know? Internal states mainly reflect knowledge recall rather than truthfulness. 2025. arXiv:2510.09033.

\bibitem[Ding and Zhang(2026)]{tonguetie2605}
Z.~Ding and X.-P.~Zhang. Causal tongue-tie: LLMs can encode causal direction, but their yes/no outputs fail to express. 2026. arXiv:2605.25891.

\bibitem[Duan(2026)]{staleness2606}
E.~Duan. Do activation monitors survive model updates? Benchmarking, predicting, and repairing
activation-monitor staleness. 2026. arXiv:2606.15980.

\bibitem[El-Shangiti et al.(2026)]{vlmcount2607}
A.~O.~El-Shangiti, A.~Nurgazy, H.~AlQuabeh, N.~Rozanov, and K.~Inui. The count is there, but misaligned:
understanding and correcting counting failures in VLMs. 2026. arXiv:2607.09544.

\bibitem[Fadnavis et al.(2026)]{deconly2605}
S.~Fadnavis, P.~Kanakaraj, and F.~Wyss. When and how long? The readout--mediator angle in temporal reasoning. 2026. arXiv:2605.29126.

\bibitem[Feng and Steinhardt(2024)]{binding2310}
J.~Feng and J.~Steinhardt. How do language models bind entities in context? \emph{ICLR}, 2024. arXiv:2310.17191.

\bibitem[Farquhar et al.(2023)]{farquhar2312}
S.~Farquhar, V.~Varma, Z.~Kenton, J.~Gasteiger, V.~Mikulik, and R.~Shah. Challenges with unsupervised LLM knowledge discovery. 2023. arXiv:2312.10029.

\bibitem[Gekhman et al.(2025)]{insideout2503}
Z.~Gekhman, E.~Ben~David, H.~Orgad, E.~Ofek, Y.~Belinkov, I.~Szpektor, J.~Herzig, and R.~Reichart. Inside-out: hidden factual knowledge in LLMs. \emph{COLM}, 2025. arXiv:2503.15299.

\bibitem[Gomezjurado Gonzalez(2026)]{repout2604}
L.~Gomezjurado Gonzalez. The long delay to arithmetic generalization: when learned representations outrun behavior. 2026. arXiv:2604.13082.

\bibitem[Gurnee et al.(2026)]{gurnee2607}
W.~Gurnee, N.~Sofroniew, A.~Pearce, et al. Verbalizable representations form a global workspace in language models. 2026. \url{https://transformer-circuits.pub/2026/workspace/}.

\bibitem[Howe(2026)]{priors2607}
G.~L.~Howe. Structure-specific representational priors causally control the grokking delay. 2026.
arXiv:2607.04333.

\bibitem[Karim et al.(2026)]{attndef2602}
A.~Karim, F.~Sheaib, Z.~Khamis, M.~Chlon, J.~Awada, and L.~Chlon. Attention deficits in language models: causal explanations for procedural hallucinations. 2026. arXiv:2602.19239.

\bibitem[Kwon(2026)]{kwon2607}
A.~Kwon. They infer what you meant: models represent communicative intent more reliably than they act on it.
2026. arXiv:2607.03598.

\bibitem[Lepori et al.(2026)]{lepori2602}
M.~A.~Lepori, T.~Linzen, A.~Yuan, and K.~Filippova. Language models struggle to use representations learned
in-context. 2026. arXiv:2602.04212.

\bibitem[Mallen et al.(2023)]{mallen2312}
A.~Mallen, M.~Brumley, J.~Kharchenko, and N.~Belrose. Eliciting latent knowledge from quirky language models. 2023. arXiv:2312.01037.

\bibitem[Li and Sreedhar(2026)]{ungrok2606}
J.~Li and D.~Sreedhar. Natural ungrokking: asymmetric control of which rules survive pretraining. \emph{ICML
FoGen Workshop}, 2026. arXiv:2606.26050.

\bibitem[Mao et al.(2026)]{agentmem2605}
X.~Mao, J.~Zhao, G.~Penn, and C.~Wang. What happens inside agent memory? Circuit analysis from emergence to diagnosis. 2026. arXiv:2605.03354.

\bibitem[Orgad et al.(2025)]{orgad2410}
H.~Orgad, M.~Toker, Z.~Gekhman, R.~Reichart, I.~Szpektor, H.~Kotek, and Y.~Belinkov. LLMs know more than they show: on the intrinsic representation of LLM hallucinations. \emph{ICLR}, 2025. arXiv:2410.02707.

\bibitem[Ravulapalli et al.(2026)]{legible2609}
M.~V.~S. Ravulapalli, S.~S. Chadha, and A.~M. Hari. Legible failures: detecting and repairing in-context binding errors. 2026. arXiv:2609.11216.

\bibitem[Sharma et al.(2026)]{decgap2604}
A.~Sharma, C.~Dawes, and S.~Raval. Dissociating decodability and causal use in bracket-sequence transformers. 2026. arXiv:2604.22128.

\bibitem[Schmied et al.(2026)]{schmied2504}
T.~Schmied, J.~Bornschein, J.~Grau-Moya, M.~Wulfmeier, and R.~Pascanu. LLMs are greedy agents: effects of RL fine-tuning on decision-making abilities. \emph{ICLR}, 2026. arXiv:2504.16078.

\bibitem[Sun et al.(2025)]{arith2507}
Y.~Sun, A.~Stolfo, and M.~Sachan. Probing for arithmetic errors in language models. 2025. arXiv:2507.12379.

\bibitem[Venkatesh(2026)]{counting2605}
S.~Venkatesh. Repeated-token counting reveals a dissociation between representations and outputs. 2026.
arXiv:2605.09239.

\bibitem[Wang and Zhu(2026)]{detext2604}
H.~Wang and M.~Zhu. The detection--extraction gap: models know the answer before they can say it. 2026. arXiv:2604.06613.

\bibitem[Xu(2026)]{devscale2605}
W.~Xu. Scale determines whether language models organize representation geometry for prediction. 2026. arXiv:2605.17084.

\bibitem[Zhao et al.(2026)]{transient2606}
Y.~Zhao, I.~Papadimitriou, and C.~Thrampoulidis. Structure before collapse: transient semantic geometry in
next-token prediction. 2026. arXiv:2606.26749.

\end{thebibliography}

\clearpage
\appendix

\section{Supporting figure}\label{app:figs}
\begin{figure}[H]
  \centering
  \includegraphics[width=0.92\linewidth]{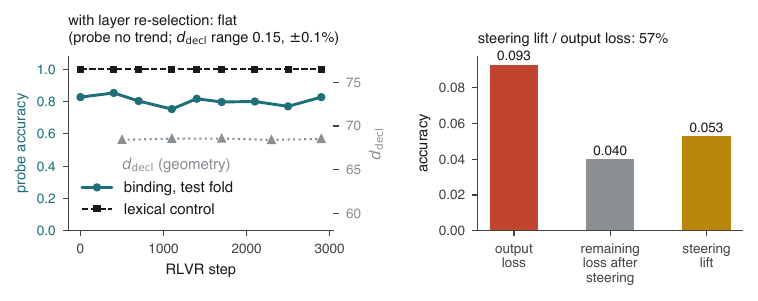}
  \caption{\textbf{RL supplement} (companion to Figure~\ref{fig:clocks}). \emph{Left}: across sampled GRPO steps,
  the refitted probe shows no detected trend. \emph{Right}: probe-guided steering recovers a portion of the measured output-accuracy loss. This intervention does not identify the cause of that loss.}
  \label{fig:rlsupp}
\end{figure}

\section{Compute resources}\label{app:compute}
All experiments ran on a single node with 8$\times$NVIDIA H100-80GB plus a CPU control-plane host. This paper
trains nothing: every measurement (probing, steering, and the per-checkpoint
clocks) is forward/backward passes over \emph{frozen} public checkpoints, and wall-clock is dominated by weight
download rather than compute. We report wall-clock rather than instrumented GPU-hours.

\section{Code and data availability}\label{app:codedata}
An earlier analysis release is public at
\url{https://github.com/RavulapalliManas/paper-outgrowing-the-readout}. This revised source package and its corrected figures are not yet in that repository. Most headline numbers are generated programmatically from released results files. A build script emits them as \LaTeX{} macros, and a verification script checks them against their sources. Summary tables derived from per-seed raw data each record a \texttt{\_provenance} field and reproduce byte-for-byte. The remaining provenance gaps are as follows. The
gpt-oss-20B origin anchor (its $\paccw$ row) and the OLMo-2 base/Instruct post-training pair are read
from a 30-model figure table whose K-sweep generator was not committed; the verification script still
checks each macro against that table, and re-running the sweep is left to future work. The
DeepSWE-v11 corpus statistics in Figure~\ref{fig:puzzle}a come from a transcribed internal summary; the underlying attempt-level records are not in the release. They serve as motivation, not evidence for the controlled binding mechanism. The new error-group, candidate-rank, and steering-arm summaries were recomputed by \texttt{analysis/p2\_saved\_artifact\_audit.py} from saved posterior, steering, and held-out logit files in the current research tree. Those inputs and the audit script are not in this arXiv source package. The
post-training stage-chain and per-step RL rows (Figure~\ref{fig:sftrlvr}) re-derive from raw batteries
in the current research tree with \texttt{analysis/agg\_posttrain.py}; the original summary file was
transcribed from a measurement log. That aggregator and those batteries are not in this arXiv source
package. The OLMo-2 developmental summary rows still lack a committed generator. The held-out decoder comparison has saved raw states, candidate logits, trial splits, and audit code; these artifacts are not yet in the public release. These gaps limit independent reproduction and must be closed before a new release claims complete source closure.

\section{Broader impacts}\label{app:impact}
The post-training results show that a probe trained at one checkpoint can lose accuracy after model updates. A monitor based on that probe needs re-validation against output behavior.
The repair machinery
could in principle steer a model toward an attacker-chosen attribute, but it self-gates and needs white-box
access to a frozen model, so it does not expand what an adversary with weights can already do.

\end{document}